\documentclass{article}

\usepackage{amssymb,latexsym}

\usepackage[pdftex]{graphicx}

\usepackage{hyperref}
\usepackage{xcolor}
\hypersetup{
    colorlinks,
    linkcolor={red!50!black},
    citecolor={blue!50!black},
    urlcolor={blue!80!black}
}

\begin{document}

\pagestyle{plain}

\newtheorem{theorem}{Theorem}[section]

\newtheorem{proposition}[theorem]{Proposition}

\newtheorem{lemma}[theorem]{Lemma}

\newtheorem{corollary}[theorem]{Corollary}

\newtheorem{definition}[theorem]{Definition}

\newtheorem{remark}[theorem]{Remark}

\newtheorem{exempl}{Example}[section]

\newenvironment{example}{\begin{exempl}  \em}{\hfill $\square$

\end{exempl}}  \vspace{.5cm}

\renewcommand{\contentsname}{ }

\title{Artificial chemistry experiments with chemlambda, lambda calculus, interaction combinators}

\author{Marius Buliga \\ 
\\
Institute of Mathematics, Romanian Academy \\
P.O. BOX 1-764, RO 014700\\
Bucure\c sti, Romania\\ 
{\footnotesize Marius.Buliga@imar.ro , mbuliga@protonmail.ch}}  \vspace{.5cm}

\date{31.03.2020}

\maketitle

\begin{abstract}
Given a graph rewrite system, a graph G is a quine graph if it has a non-void maximal  collection of non-conflicting matches of left patterns of graphs rewrites,  such that after the  parallel application of the rewrites we obtain a graph isomorphic with G. Such graphs exhibit a metabolism, they can multiply or they can die, when reduced by a random rewriting algorithm.

These are introductory notes to the pages of artificial chemistry experiments with chemlambda, lambda calculus or interaction combinators, available from the entry page \href{https://chemlambda.github.io/index.html}{chemlambda.github.io} \cite{entry}. The experiments are bundled into pages, all of them based on a library of programs, on a database which contains hundreds of graphs and on a database of about 150 pages of text comments and  a collection of more than 200 animations, most of them which can be re-done live, via the programs. There are links to public repositories of other contributors to these experiments, with versions of these programs in python, haskell, awk or javascript. 
\end{abstract}

\section{How not to read these notes}

These  notes serve to clarify the basics and as a reference for the concepts used in the programs and pages of the 
\href{https://chemlambda.github.io/index.html}{library of experiments} \cite{entry}. You only need a browser with the javascript turned on to start exploring these artificial chemistry experiments. 

But if you don't understand what are you looking at, then come back to these notes. 

\section{How you can contribute}

\paragraph{Find new quines.} By going to the page \href{https://mbuliga.github.io/quinegraphs/quinecheck.html}{How to test a quine}, read the instructions. After this choose from the menu one of the variants which create random graphs, like \href{https://mbuliga.github.io/quinegraphs/quinecheck.html#random_egg_A_L_FI_FO}{this} or \href{https://mbuliga.github.io/quinegraphs/quinecheck.html}{this} for chemlambda, or \href{https://mbuliga.github.io/quinegraphs/quinecheck.html#random_egg_G_G_D_D}{this} for interaction combinators. Copy the CODE and mail it to me. If the quine is new then congratulations, you can name this new life form.

\paragraph{Create a new chemistry.} For this you have to read these notes. The programs are commented too. A new chemistry needs new nodes and new graph rewrites. The nodes are defined in the script \href{https://github.com/mbuliga/quinegraphs/blob/master/js/nodes.js}{nodes.js}, as explained in section \ref{mol-nodes-prog}. The chemistries are defined in the script \href{https://github.com/mbuliga/quinegraphs/blob/master/js/chemistry.js}{chemistry.js}, as explained in section \ref{rewrites-prog}. You can use \href{https://mbuliga.github.io/quinegraphs/quinelab.html}{The quine lab} to input new graphs, in mol format. Or you can clone or download the \href{https://github.com/mbuliga/quinegraphs}{quinegraphs} Github repository, then edit the chemistry.js, nodes.js to add a new chemistry, then edit the \href{https://github.com/mbuliga/quinegraphs/blob/master/js/iceMol.js}{library of graphs} to add new graphs. You can add comments for your new graphs in \href{https://github.com/mbuliga/quinegraphs/blob/master/js/iceComments.js}{comments}. I would like to see, for example, new chemistries like one for the project \href{https://github.com/moonad/Formality}{Formality} or to invent a chemistry which would be able to graphically parse lambda terms as an enhancement of the  \href{https://mbuliga.github.io/quinegraphs/lambda2mol.html}{lambda calculus to chemlambda parser}. 

\paragraph{Make artificial life animations.} The \href{https://chemlambda.github.io/collection.html}{Chemlambda collection of animations} contains hundreds of recordings of artificial chemistry simulations. You can remake such simulations and record the screen. There is only one rule: minimal editorial intervention. You can speed the screencast to fit into a short time and you have the freedom to chose the framing of the animation. In \href{https://github.com/mbuliga/quinegraphs/blob/master/js/parameters.js}{parameters.js} you can modify the dimension of the simulations window, the gravity, the forces acting on the links of the graphs, or the color of the nodes. During the simulation, you can translate the graph by mouse drag, or you can zoom with the mouse wheel. The screencast can be then transformed into an animated gif, as an art creation. 

\paragraph{Find bugs, suggest corrections.} Please inform me about research lines which are not mentioned here, but are relevant for the subject. I am a mathematician with a geometry and applied mathematics background, therefore help me to not reinvent the wheel too often. If you like programming, then contribute to one of the repositories or contact me. 

\paragraph{Ask questions.} A good question is a great contribution. These experiments are the outcome of a research subject which has ramifications in fields which seem far away from artificial chemistry, like metric geometry. Maybe your research involves applications of graph rewriting in a new context. 

\section{Introduction}

The experiments are bundled into pages, all of them based on a library of programs, on a database which contains hundreds of graphs and on a database of about 150 pages of text comments and  a collection of more than 200 animations, most of them which can be re-done live, via the programs. 

The \href{https://chemlambda.github.io/index.html}{entry page of the library} \cite{entry} contains links to the following pages: 
\begin{enumerate}
\item[-] \href{https://mbuliga.github.io/quinegraphs/ice.html}{Interaction combinators and chemlambda quine graphs}, to explore various quine graphs from chemlambda and interaction combinators \cite{lafont-comb} and to generate new ones. Given a graph rewrite system, a graph G is a quine graph if it has a non-void maximal  collection of non-conflicting matches of left patterns of graphs rewrites,  such that after the  parallel application of the rewrites we obtain a graph isomorphic with G. Such graphs exhibit a metabolism, they can multiply or they can die, when reduced by a random rewriting algorithm.
\item[-] \href{https://mbuliga.github.io/quinegraphs/quinecheck.html}{How to test a quine}, which allows you to check or to prove that a graph is indeed a quine graph or not
\item[-] \href{https://mbuliga.github.io/quinegraphs/quinelab.html}{The quine lab} allows you to input or output an interesting graph or it's reductions, with examples to play with. The mol format used to encode the graphs is a variant of linear graphs \cite{bawden1} \cite{bawden2}, or port graphs \cite{stewart} familiar to those interested in interaction nets \cite{lafont-nets} and linear logic \cite{girard-linear}.
\end{enumerate} 
For those interested in lambda calculus \cite{church} there are two pages: 
\begin{enumerate}
\item[-] \href{https://mbuliga.github.io/quinegraphs/lambda2mol.html}{Lambda calculus to chemlambda parser}, to transform untyped lambda terms into chemlambda molecules, to reduce them graphically and then to output the result,
\item[-] \href{https://mbuliga.github.io/quinegraphs/ouroboros.html}{The Ouroboros} explains how the first chemlambda quine graph, the ouroboros, was discovered from the graphical reduction of a lambda term.
\end{enumerate}
Finally, for those interested into the possibilities of these artificial chemistry, or into artificial life, there is a big collection of animations, with a more artistical nature. These animation are produced from simulations performed with the programs from the library, with a minimal editing concerning only the speed and the framing of the visual output. They come with abundant comments.
\begin{enumerate}
\item[-] \href{https://chemlambda.github.io/collection.html}{Chemlambda collection of animations}
\item[-] \href{http://imar.ro/~mbuliga/collection.html}{Same collection with bigger animations}
\end{enumerate}
was produced initially as a Google+ collection, after an effort of public research communication. It is now enhanced by the possibility to re-done the animations in javascript, with the programs from the library. 

The entry page contains links to more informations and, more importantly, to public repositories of other contributors to this experiment. You find there versions of these programs in python, haskell or my original programs in awk. I want to express my thanks for all contributions. This version of the programs, in javascript, is actually based on the first javascript-only port of my original programs. All the visualizations are done via the excellent d3.js. 

\tableofcontents

\section{About this project}

The source of these experiments is the discovery of a graphical formalism for differential calculus in metric spaces known as sub-riemannian spaces. See \cite{buliga1} \cite{buliga2} \cite{buliga3}, if you are more interested in geometry or analysis in metric spaces, or 
\cite{buliga4} for the most recent mix of lambda calculus with emergent algebras. 

This formalism of dilation structures, or emergent algebras,  composes graph rewriting with a passage to the limit to prove differential or algebraic statements. Trying to understand the computational power of this formalism I started from an algebraic point. Emergent algebras are families of idempotent right quasigroups, indexed by a topological commutative groups, along with a topological structure which allows uniform limits. This algebraic framework did not allow to understand the full generality of the graphical formalism. Therefore I changed to an interaction nets like proposal called graphic lambda calculus \cite{buligaglc}. As my intuition was that these graphs and graph rewrites represent a sort of programs with space itself, it follows that the graphical reduction algorithm should be the most simple one. This is how I concentrated on a purely local version of graphic lambda calculus,  an artificial chemistry called chemlambda (from "chemistry" and "lambda calculus") which sees graphical rewrites as chemical reactions mediated by enzymes (local machines). Here artificial chemistry is used not in the usual sense \cite{banzhaf}, \cite{dittrich} but in the sense of asynchronous graph rewrite automata \cite{tomita}. 

Chemlambda artificial chemistry was initially proposed with the name chemical concrete machine \cite{buligachem} (an allusion to Berry and Boudol chemical abstract machine \cite{cham}).  A version  of the graph rewrites, but without the algorithm of applications, appears with the name chemlambda in a collaboration with Kauffman \cite{chemlambda1}, where we also explore interactions of knot theory with computation. Knot theory is related to this subject via the fact that idempotent right quasigroups which are also distributive are quandles, or quandles are algebraic structures which encode the Reidemeister moves (graph rewrites) of knot diagrams. Kauffman \cite{kauffman1} \cite{kauffman2} \cite{kauffman3} has a long time interest in the relation between knot theory and computation. Previously with Kauffman, we proposed a model of computation based on graphic lambda calculus and actors in \cite{glc}, which can be recognized as a mathematicians rediscovery of a version of Bawden \cite{bawden1} \cite{bawden2} linear graphs applications in distributed computing. 

The actual version of chemlambda as an artificial chemistry is built into the first collection of programs and the associated demo pages, referenced in this library. These programs appeared from  the realization that, besides the mathematician specialized interest, there may be applications of these ideas in chemistry. Lambda calculus has been used with a chemical theme, by Fontana and Buss \cite{fontana1} 
\cite{fontana2}, in their Alchemy (algorithmic chemistry). They consider populations of lambda terms, whose members interact by lambda calculus application. Graph rewriting is of obvious interest for chemistry applications and frameworks like the Graph Grammar Library \cite{flamm}, which allow creation of graph rewriting as chemical reactions rules, existed. The new idea brought by chemlambda was that interaction nets like graph rewriting could be the base of interesting artificial chemistries, or why not, real chemistry. Hence the idea of a molecular computer 
\cite{molecular} which is simply a (chemical) molecule (in this toy chemistry) which computes via random chemical rewrites, that is independently, not in a controlled medium or laboratory. The main question was if interesting computations can be done, taking as inspiration graphical reductions of lambda calculus terms, in a interaction nets asynchronous automaton. (Chemlambda however, differently from interaction nets, does have conflicting rewrites.)

The chemlambda project took a completely open turn, based on programs and demonstrations which could be validated by reproduction or by porting of the programs into other languages (the first programs used awk for the processing and d3.js for visualization). And indeed the project was validated by the other contributors which are listed in the library entry page.

\section{Informal description of patterns, mols, graph rewrites}

This is an informal description of a variant of linear graphs \cite{bawden1} \cite{bawden2}, or port graphs \cite{stewart} and their double push-out \cite{dpo} graph rewriting, as they appear in the artificial chemistry experiments described in these notes. 

We are concerned with graphs which have a finite number of nodes. Nodes are connected with edges via node ports. The valence of a node is the number of its node ports. Each edge connects two different node ports. Every node port is connected to an edge. Each node has a type from the list $NT$ of node types. The type $ A \in NT$ of a node gives the valence $ar(A)$ of the node, as well as a numbering of the node ports (from 0 to $ar(A) - 1$ in the programs, or from 1 to ar(A) in the description). 

Any graph made of a finite number of nodes, each node of a type from the list of node types, admits a mol notation. We need an infinite alphabet $Edge$ of symbols which we use to decorate the edges of the graph, such that every two different edges have different decorations. We arbitrarily order the nodes of the graph. Then we produce the mol notation of the graph which is a list of node items, one for each node. Each item is a list which starts with the node type, then with the edge decorations of each edge incident to a node port, in the order of the node ports. 

For example the list: 
$$((A,a,b,c), (A,b,d,e), (B,c,a,d), (A,e,f,f))$$
describes a graph with 4 nodes, 3 of them of type A and one of type B. All nodes have valence 3. There are 6 edges, decorated with the symbols a,b,c,d,e,f. (By using the order of the nodes given by the list) we know that the edge "a" connects the 1st port of the first node with the 2nd port of the second node, and so on until the last edge, "f", which connects the 2nd and 3rd ports of the 4th node. 

Further we shall write node items and node words  in the following format: one node item per line, without parantheses or commas, like this
$$\begin{array}{llll}
    A   &    a   &     b    &     c  \\
    A   &    b   &     d    &     e  \\
    B   &    c   &     a    &     d  \\
    A   &    e   &     f    &     f        
\end{array} 
$$
Notice that each edge decoration appears exactly twice.

The mol notation of a graph is unique up to the renaming of the edges (decorations) and up to  permutation of the node items. Similarly, two graphs are isomorphic if they admit the same mol notation, again up to renaming of the edges and up to permutation of the node items. 

A pattern is a collection of nodes of the graph, together with the internal edges (joining the nodes of the pattern) and with their external half-edges, but seen in the same mol notation. For the example of a graph taken previously, any collection of lines in the mol notation is a pattern, like this one: 
$$\begin{array}{llll}
    A   &    a   &     b    &     c  \\
    B   &    c   &     a    &     d        
\end{array} 
$$
In the mol notation of a pattern, each edge decoration appears at most twice. Those edge decorations which appear once are the external half-edges. 

A match of a pattern into a graph (described by its mol notation) is a pair of functions $(f,g)$, where $g$ is an injective function from the nodes (lines in the list) of the pattern to the nodes of the graph, and $f$ is a renaming of the edge decorations, such that on those those decorations which appear twice in the pattern the renaming is injective, and on the external half-edges  the renaming is at most 2-to-1. 

For example if we take the pattern:
$$\begin{array}{llll}
    A   &    1   &     2    &     3  \\
    A   &    3   &     4    &     5        
\end{array} 
$$
then there is a match with the graph taken as an example: 
$$\begin{array}{llll}
    A   &    1   &     2    &     3  \\
    A   &    3   &     4    &     5        
\end{array} \hspace{.5cm} \rightarrow \hspace{.5cm}
\begin{array}{llll}
    A   &    a   &     b    &     c  \\
    \mathbf{A}   &    \mathbf{b}   &     \mathbf{d}    &     \mathbf{e}  \\
    B   &    c   &     a    &     d  \\
    \mathbf{A}   &    \mathbf{e}   &     \mathbf{f}    &     \mathbf{f}        
\end{array} 
$$
Indeed, here the function $g$ matches the 1st line of the pattern with the 2nd line of the mol notation of the graph and the 2nd line of the pattern with the 4th line of the mol notation of the graph. The function $f$ is defined as: 
$$f(1) = b \, . \, f(2) = d \, f(3) = e \,  f(4) = f(5) = f$$

A graph rewrite $\displaystyle LHS \rightarrow RHS$ is defined by two patterns with the same external half-edges, called the LHS and the RHS patterns of the rewrite.  For our example suppose that we have the following graph rewrite: 
$$\begin{array}{llll}
    A   &    1   &     2    &     3  \\
    A   &    3   &     4    &     5        
\end{array} \hspace{.5cm} \rightarrow \hspace{.5cm}
\begin{array}{llll}
    C   &    1   &     2    &    \\
    C   &    4   &     5    &            
\end{array}
$$
In the usual style of \href{http://eprints.adm.unipi.it/1950/1/TR-96-17.ps.gz}{DPO} rewrites \cite{dpo} (adapted to the mol notation) the application of the graph rewrite is a 3 parts process, where: 
\begin{enumerate}
\item[-] we start from a match of the LHS pattern into the graph, for example the one considered previously,
\item[-] we produce an intermediary mol notation by deleting the lines in the mol notation of the graph which are in the scope of the match thus obtaining 
$$
\begin{array}{llll}
    A   &    a   &     b    &     c  \\
    B   &    c   &     a    &     d  \\
\end{array} 
$$
\item[-] we concatenate the intermediary mol notation with a copy of the RHS pattern, such that all the internal edges have fresh decorations (not the case in our example, where the RHS does not have internal edges) and such that the external edges inherit the decoration from the match of the LHS pattern. In our example we get:
$$
\begin{array}{llll}
    A   &    a   &     b    &     c  \\
    B   &    c   &     a    &     d  \\
    \mathbf{C}   &    \mathbf{b}   &     \mathbf{d}    &    \\
    \mathbf{C}   &    \mathbf{f}   &     \mathbf{f}    &      
\end{array} 
$$
\end{enumerate}

In order to allow graphs with some free edges we suppose that we always have  in $NT$ 1-valent nodes named "FR*" where "*" is a word of our choice. For example a pattern which has external half-edges, like 
$$\begin{array}{llll}
    A   &    a   &     b    &     c  \\
    A   &    c   &     d    &     e        
\end{array} 
$$
could be seen as the mol notation of a graph with free half-edges a, b, d, e, by adding to it 4 FR* nodes, like this: 
$$\begin{array}{llll}
    A   &    a   &     b    &     c  \\
    A   &    c   &     d    &     e  \\
    FRIN & a     &          &        \\
    FRIN & b     &          &        \\
    FROUT & c     &          &        \\
    FROUT & d     &          &        \\
\end{array} 
$$
so that now each edge decoration appears exactly twice. This completion of a pattern to a mol notation does not erase the difference between mols and patterns though. 

There are two kinds of graph rewrites which we want to allow. We want  to allow parallel graph rewrites which don't share nodes, but they do share half-edges. Also we want to allow graph rewrites which delete nodes. That is why we suppose that we always have in $NT$ 2-valent nodes named "Arrow*". Moreover we shall always add to the graph rewrite systems considered a family of graph rewrites called "COMB", which consist into the elimination of the Arrow* elements. For example suppose we have a graph rewrite which deletes a pair of neighboring nodes and joins the dangling half-edges. This can be done by a graph rewrite of the form
$$\begin{array}{llll}
    A   &    1   &     2    &     3  \\
    B   &    3   &     4    &     5        
\end{array} \hspace{.5cm} \rightarrow \hspace{.5cm}
\begin{array}{llll}
    Arrow   &    1   &     5    &    \\
    Arrow   &    4   &     2    &            
\end{array}
$$
This allows to apply several such rewrites in parallel, without confusion concerning the wiring of the edges. After the rewrites are done (one or more in parallel), we end up with the following two situations. A node connected to an Arrow*, or an Arrow* with both ports connected to the same edge. The COMB rewrites are then used to erase the Arrow* elements.  For any $A \in NT$ 
$$\begin{array}{llll}
    A   &    ...   &    e_{i}   &  ...  \\
    Arrow*   &    e_{i}   &     b  &       
\end{array}  \hspace{.5cm} \rightarrow  \hspace{.5cm} 
\begin{array}{llll}
    A   &    ...   &    b   &  ...  
\end{array}  
$$
and 
$$\begin{array}{llll}
    Arrow*   &    a  &    a   &          
\end{array} \, \rightarrow \, 
\emptyset
$$ 

\section{Mathematical description of patterns, mols, graph rewrite systems}

Linear graphs \cite{bawden1} \cite{bawden2}, or port graphs \cite{stewart} and their double push-out \cite{dpo} graph rewriting, appear mostly in relation to interaction nets \cite{lafont-nets} and linear logic \cite{girard-linear}. Here we present a variant of this formalism, with a chemical blend, as it used in these experiments. 

\subsection{Notations}

A vector $\displaystyle v = \left[a_{1},...,a_{p} \right]$  is a function $\displaystyle v: \left\{1,...,p\right\} \rightarrow M$, $\displaystyle v(k) = a_{k}$. The support $\displaystyle supp(v)$ of the vector $v$ is the image of the function $v$, that is the set of values $\displaystyle a_{k}$. The arity of the vector is $\displaystyle ar(v) = p$.

\subsection{Molecules}

Let $NT$ be a finite, non-empty set of node types or colors and node type arity a function: 
$$nar: NT \longrightarrow \mathbb{N}^{*}$$

A $NT$-molecule pattern is a triple of sets $M = (H,N,E)$, where $N$ is a $NT$-colored partition of $H$ into ordered subsets, called nodes, and $E$ is a set of unordered pairs of half-edges, called edges. 

\begin{definition}
A $NT$-molecule pattern is a triple of sets $M = (H,N,E)$, where:

(a) $H$ is a finite, non-empty set of half-edges,

(b) $N$ is a set of nodes. Each node $\displaystyle n \in N$ 
$$n = \left[c, h_{1}, ..., h_{nar(c)} \right] $$
is colored with a  $color(n) = c \in NT$  and has a vector $\displaystyle half(n) = \left[h_{1}, ..., h_{nar(c)} \right]$  of half-edges incident to the node, of arity $nar(c)$,

(c) $E$ is a set of edges. Each edge $\displaystyle e \in E$ is a set of two different half-edges incident to the edge, \\ \\
which satisfy the conditions: \\

(d) for any half-edge $\displaystyle h \in H$ there exists and are unique a node $\displaystyle n = center(h) \in N$ and an index $\displaystyle k = port(h) \in \left\{1,...,nar(color(n))\right\}$ such that $\displaystyle h = half(n)_{k}$,

(e) for any half-edge $\displaystyle h \in H$ there is at most one edge $\displaystyle e = edge(h) \in E$ such that $h \in e$. \\ \\
The half-edge $\displaystyle h \in H$ is bound in $M$ if there exists an edge $\displaystyle e = edge(h) \in E$ as in condition (e). We write $\displaystyle h \in Bound(M)$. 
 Otherwise we say that $h$ is free in $M$, $\displaystyle h \in Free(M)$, and we define $\displaystyle edge(h) = h \in Free(M)$. 

A molecule is a molecule pattern without free half-edges. 

For completeness we accept as a molecule the empty one $\displaystyle \emptyset$, with no half-edges. 
\label{molecule-math}
\end{definition}

We may supplement the node type arity with a node valence function, which associates to each node type $c \in NT$ a vector 
$$nV(c) = [v_{1}, ..., v_{nar(c)}]$$ 
with support in $\left\{0,1\right\}$. We say that the port $k$ of the color $c$ is "in" if $\displaystyle nV(c)_{k} = 0$, otherwise is "out".

An oriented molecule pattern $M = (H,N,E)$ is a molecule as in definition \ref{molecule-math}, with the following modifications: edges are ordered, they have a source and a target and sources of edges are connected to "out" ports of nodes and targets of edges are connected to "in" ports of nodes. 
More precisely, we have the following modifications of conditions (c) and (e): \\ \\ 
(c) $E$ is a set of edges. Each edge $\displaystyle e \in E$ is a pair $\displaystyle (source(e), target(e)) \in H^{2}$ 
with $\displaystyle source(e) \not= target(e)$, \\ \\ 
(e) for any half-edge $\displaystyle h \in H$ there is at most one edge $\displaystyle e = edge(h) \in E$ such that 
$\displaystyle h = source(e)$ or $\displaystyle h = target(e)$. Let $\displaystyle n = center(h) \in N$ be the node incident with $h$ and 
 $\displaystyle k = port(h)$ be the index of $h$ in $n$. Let $\displaystyle c = color(n)$ be the color of the node.   If $\displaystyle h = source(e)$ then $\displaystyle nV(c)_{k} = 1$.  If $\displaystyle h = target(e)$ then $\displaystyle nV(c)_{k} = 0$. 

From definition \ref{molecule-math} we see that a molecule pattern is equally described by the associated functions $color$, $half$, $port$, $center$ and $edge$. A molecule pattern morphism is one which commutes with all these functions. 

\begin{definition}
A morphism from a $NT$-molecule pattern  $M = (H,N,E)$, with associated $color$, $half$, $port$, $center$ and $edge$, to another $NT$-molecule pattern  $M' = (H',N',E')$, with associated $color'$, $half'$, $port$, $center'$ and $edge'$, is a function
$$F: H \rightarrow H'$$
with the properties: 

(a) for any $\displaystyle h_{1}, h_{2} \in H$, if  $\displaystyle center(h_{1}) = center(h_{2})$ then $\displaystyle center'(F(h_{1})) = center'(F(h_{2}))$ 

(b) for any $h \in H$ $\displaystyle port'(F(h)) = port(h)$, 

(c) for any $h \in H$ $\displaystyle color'(center'((F(h))) = color(center(port(h)))$, 

(d) for any $h \in Bound(M)$ we have $F(h) \in Bound(M')$ and $$\displaystyle edge'(F(h)) = \left\{ F(g) | g \in edge(h) \right\}$$ 

Let $(NT, nar)$ and $(NT',nar')$ be two sets of node types with associated arity functions. A connection morphism from a $NT$-molecule pattern  $M = (H,N,E)$, with associated $color$, $half$, $port$, $center$ and $edge$, to another $NT'$-molecule pattern  $M' = (H',N',E')$ is a pair $(K,F)$ 
$$K: NT \rightarrow NT' \quad , \quad F: H \rightarrow H'$$
such that $F$ satisfies conditions (a), (b), (d) and $F$, $K$ satisfy the modified condition: 

(c') for any $c \in NT$ $nar'(K(c)) = nar(c)$ and for any $h \in H$ $$\displaystyle color'(center'((F(h))) = K(color(center(port(h))))$$ 

For oriented molecule patterns there are similar definitions of morphisms and connection morphisms. 
\label{molmorphism-math}
\end{definition}

Notice that a morphism does not preserve free half-edges: it is not true, in general, that if $h \in Free(M)$ then $F(h) \in Free(M')$. Indeed, it is possible for the images of two free half-edges in $M$ to form an edge in $M'$. 

If we accept among the node types 1-valent node types called FREE, or FRIN (free in) or FROUT (free out) in the case of oriented molecule patterns, then we can transform any molecule pattern $M$ with $Free(M) \not= \emptyset$ into a molecule $Cap(M)$ simply by adding for any 
half-edge $h \in Free(M)$ a new half-edge $h'$, a node $\displaystyle [FREE, h']$, (or $\displaystyle [FRIN, h']$ if $h$ is "in" or 
$\displaystyle [FROUT, h']$ $h$ is "out"), and an edge $\displaystyle \left\{h,h'\right\}$ (or $\displaystyle [h',h]$ if $h$ is "in" or 
$\displaystyle [h,h']$ if $h$ is "out"). 

Remark that $\displaystyle M \mapsto Cap(M)$ does not induce a functor in the category of mol patterns, because morphisms do not preserve free half-edges. However, if $M$ is isomorphic with $M'$ then $Cap(M)$ is isomorphic with $Cap(M')$. 

\subsection{The graph associated to a molecule}

Let $C$ be a finite, non-empty set of colors and $\displaystyle nar: C \rightarrow \mathbb{N}^{*}$ be a color arity function.

\begin{definition}
A $C$-colored graph is a triple $(N,E,color)$ where $N$ is a finite nonempty set of nodes, $color: N \rightarrow C$ and E is a set of edges. Each edge is a set of two different  nodes. 

If a node belongs to an edge then we say that the edge is incident to the node. If two different nodes belong to an edge then we say that the edge links the nodes. 

The arity $ar(n)$ of a node $n$ is the number of edges which are incident to that node. We demand that: 
$$ar(n) = nar(color(n))$$
for any node of the graph. 
\label{coloredgraph-math}
\end{definition}

To any molecule $M$ is associated a colored graph $Graph(M) = (Nodes, Edges)$. 

Denote by $m$ the maximal arity of nodes in the molecule $M$. Then we add to the set of colors the set $\displaystyle \left\{1,...,m\right\}$ (we suppose of course that $C$ is disjoint from this set) and we obtain a new set of colors $C'$. We exetnd the node type arity 
$$nar: C \rightarrow \mathbb{N}^{*}$$
to a color arity (denoted the same) defined over $C'$, with 
$$nar(k) = 2$$
for each $\displaystyle k \in \left\{1,...,m\right\}$. 

To each node  $\displaystyle n = [c,h_{1},...,h_{p}]$ in $M$, of arity p, we add p + 1 nodes  in $Nodes$, the $k$-th node colored with $k$, for $k = 1,...,p$, and the last node colored with $c = color(n)$. This last node is named a center node and the other nodes are named port nodes. We add edges in $Edges$ from the center node to each of its port nodes, we call these internal edges. Finally, for each pair of half-edges which form an edge in $M$, we add an external edge linking the corresponding node ports. 

If $M$ is a molecule pattern with $Free(M) \not= \emptyset$, then by definition $$Graph(M) = Graph(Cap(M))$$

$Graph(M)$ determines the molecule pattern $M$ up to isomorphism. In general a $M \mapsto Graph(M)$ is a functor only from the category of molecules to the category of graphs, because, again, morphisms of pattern molecules do not preserve free half-edges. 

\subsection{Mol notation}
The mol notation corresponds to Bawden \cite{bawden2} linear graph notation. 

\begin{definition}
An $NT$-mol node with edge tags in the set $Tag$ is a vector 
$$\bar{n} = \left[c, tag_{1},..., tag_{nar(c)}\right]$$
where $\displaystyle c = color(\bar{n}) \in NT$ is the node type of the mol node and $\displaystyle half(\bar{n}) = \left[tag_{1}, ..., tag_{nar(c)} \right]$ is a vector of tags, elements of $Tag$, such that any $tag \in Tag$ occurs at most twice. 

An $NT$-mol pattern with edge tags in the set $Tag$ is a vector of $NT$-mol nodes, such that any $tag \in Tag$ occurs at most twice over all mol nodes vectors. 

In the oriented version, when $NT$ comes with a node valence function $nV$, an $NT$-mol pattern with edge tags in the set $Tag$ is mol pattern  such that every tag $tag \in Tag$  which occurs twice, it does appear two positions (ports) which are "in" and "out". 

In general, the tags which occur twice in the mol pattern $M$ are said to be bounded, their set is denoted $Bound(M)$. The tags which occur once we said to be free, their set is denoted $Free(M)$.  
\label{mol-abs-math}
\end{definition}

A mol pattern (or just mol) is a record of a molecule pattern. 

\begin{definition}
Let $M = (H,N,E)$ be  $NT$-molecule pattern, $Tag$ a set of tags in bijection $\displaystyle tag: E \rightarrow Tag$ with $E$, $\displaystyle Id: \left\{1,...,m\right\} \rightarrow N$ a bijective numbering of nodes. 
 
The mol (notation) of $M$ is the mol pattern $\displaystyle \left[\bar{n}_{1}, ..., \bar{n}_{m} \right]$,  where for each $\displaystyle  k \in \left\{1,...,m\right\}$, 
$$\displaystyle n = Id(k) = \displaystyle n = \left[c, h_{1},..., h_{nar(c)} \right]$$
$$\bar{n}_{k} = \left[c, tag(h_{1}),..., tag(h_{nar(c)}) \right]$$
\label{mol-math}
\end{definition}

\subsection{Local machines}
\label{localmachines}

Given a set of node types $NT$, with node type arity or node valence function in the oriented case, a set $Tag$ of edge tags, a finite set of special symbols and a natural number $K$, we define a $K$-local machine. This is a Turing machine which has two tapes: an IO (input-output tape)  and a work tape, and a set of internal states, such that the size of the work tape plus the number of possible internal states is at most $K$. 

The tapes are made by cells, each cell can contain a node type, a tag, or a special symbol. 

The IO tape has to contain at each moment only a mol pattern and special symbols, like field separators of a node vector or line separator for the mol vector. The IO tape head, after reading a cell, if it reads a tag it can then jump to the beginning of the mol node containing it,  or to the cell which contains the other occurence of the tag (if it exists), or to one of the other tags from the same mol node, or to the end of mol pattern. 

The machine can read from on the IO tape a cell, write a whole mol node at the end of the mol pattern, or delete a whole mol node. 
For each read/write or delete from/to the IO tape the machine writes the same on the work tape. The machine can read from the work tape and it can write only on the blank cells of the work tape. 

When the machine halts (by writing a halt symbol on the work tape) it either deletes the whole work tape and the IO head jumps to a random mol node beginning, or the machine halts. 

The machine may have a source of new tags (not already present on the IO tape) or not. The machine may have access to a random coin.  

Section \ref{hapax} explains why we may not need a source for new tags.

\section{Molecules for programmers}

If you are familiar with linear graphs \cite{bawden1} \cite{bawden2}, or port graphs \cite{stewart} and their double push-out \cite{dpo} graph rewriting, then you can use this section for faster understanding of the programs in the library. 

Such graphs are familiar to programmers interested in interaction nets \cite{lafont-nets} and linear logic \cite{girard-linear}. However, in these experiments, the accent is on the use of the simplest, purely local algorithms, and not on the semantics or the high level point of view. Here we want to understand if these simplest algorithms can do something interesting in the artificial chemistries we consider.

In order to define mols (molecules) and mol patterns, we introduce mol node types and 
their nodeValence vectors.

\subsection{Mol nodes}
\label{mol-nodes-prog}

The mol nodes are defined in the script \href{https://github.com/mbuliga/quinegraphs/blob/master/js/nodes.js}{nodes.js}.

\paragraph{Mol nodes types.}
Let NT be a finite  set of mol node types. In the script NT is in the vector autoFilter.

\begin{enumerate}
\item[]\begin{verbatim}
autoFilter = [
"L","A","FI","D","FOE","FOX","FO","T","Arrow","GAMMA","DELTA"
];
\end{verbatim}
\end{enumerate}
nodeValence associates to any mol node type a valence vector, with elements 0 or 1, whose length is the valence of the mol node. 

\begin{enumerate}
\item[]\begin{verbatim}
nodeValence = {
  "L":  [0,1,1], // (12) , 1-z 
  "A":  [0,0,1], // (231), (z-1)/z
  "FI": [0,0,1], // (312), 1/(1-z)
  "D": [0,0,1],  // ()   , z
  "FOE":[0,1,1], // (23) , 1/z 
  "FOX": [0,1,1], // (13) , z/(z-1)
  "FO": [0,1,1], //
  "T":  [0],
  "FRIN":[1],
  "FROUT":[0],
  "Arrow":[0,1],
// interaction combinators
  "GAMMA":[0,0,0],
  "DELTA":[0,0,0],
}
\end{verbatim}
\end{enumerate}

Mathematically, given the set NT of mol node types, the nodeValence is a function which 
associates to a mol node type "t" the word w = nodeValence(t). Here the word w is made of 
letters "0" and "1". The valence of the mol node type is the length of the word w. 

In the  definition of nodeValence we see some mol node types with comments. These mol 
node types can be decorated with permutations of 3 elements, or with elements of the anharmonic 
group. These decorations are a bridge from the present work to emergent algebras (see for 
example \href{https://arxiv.org/abs/1807.02058}{arXiv:1807.02058}) which are "commutative" 
in the sense that they satisfy a rewrite called "the shuffle trick". 

Here we are going to use this correspondence only heuristically, for example to choose the form 
of the rewrites "DIST", which increase the number of nodes. 
See more about this correspondence at: 
\begin{enumerate}
\item[-] \href{https://mbuliga.github.io/kali24.html}{anharmonic lambda calculus}
\item[-] the tool to \href{https://mbuliga.github.io/rhs.html}{choose DIST rewrites}
\item[-] the commented js script \href{https://mbuliga.github.io/rhs.js}{rhs.js}
\end{enumerate}
A detailed explanation of the relation with commutative emergent algebras will appear in the future.

\paragraph{Mol nodes.} 
Given a nonempty, finite set of edge tags E, a mol node whose ports are E-decorated is a vector 
(t, $\eta$)
where:
\begin{enumerate}
\item[-] t is a mol node type, called the type of the mol node
\item[-] $\eta$ is a word over the alphabet E, such that:
\begin{enumerate}
\item[-] any letter from E appears at most twice,
\item[-] the length of $\eta$ equals the valence of t.
\end{enumerate}
\end{enumerate}

For a mol node we shall use a notation like 
\begin{enumerate}
\item[]\begin{verbatim}
L a b c
\end{verbatim}
\end{enumerate}
where "L" is a mol node type and "abc" is the word $\eta$ with letters from an alphabet E 
which contains a, b, c. You can see that the mol node type L has nodeValence vector [0,1,1], whose length is 3, equal to the length of the word abc. 

\subsection{Mol patterns} 

A mol pattern (which is E-decorated) is a vector of mol nodes, such that any letter from E 
appears at most twice over all mol nodes. 

Any letter which appears exactly once in a mol pattern P, is called a free edge. The set of free edges of the mol pattern P is denoted by 
Free(P). The set of edges of the mol pattern P which are not free is denoted by Bound(P).

A mol (which is E-decorated), is a mol pattern without free edges. 

For mol patterns, in particular for mols, we use a notation like 
\begin{enumerate}
\item[]\begin{verbatim}
L a b c
A c d e
\end{verbatim}
\end{enumerate}
which is amenable to a record with a line separator (here we use newline) which separates the record into lines. Each line is a mol node, itself a record with a field separator (here we use 
space, therefore we can't admit space or newline in the alphabet E). 

Such records are available in \href{https://github.com/mbuliga/quinegraphs/blob/master/js/iceMol.js}{iceMol.js}, in the form of the function molLibrary(). In those records the line separator is "\^". 

Note that any mol pattern can be turned into a mol by adding some new mol nodes "FRIN" (i.e. "free in") and "FROUT" (i.e. "free out"). 

\subsection{Molecules}

A molecule is an equivalence class of mol patterns, up to the reordering of mol nodes and up to renaming of the edge decorations. This equivalence is defined in terms of morphisms of mol patterns. 

Consider two mol patterns: 
\begin{enumerate}
\item[] $\displaystyle P = \left[(t_{i}, \eta_{i})\right]_{i = 1...n}$, which is E-decorated,
\item[] $\displaystyle Q = \left[(t{'}_{j}, \eta^{'}_{j})\right]_{j = 1...m}$, which is F-decorated.
\end{enumerate}
 A morphism from P to Q is a pair of functions $\left[f,d\right]$, where $\displaystyle f:\left\{1...n\right\} \rightarrow \left\{1...m\right\}$ and $\displaystyle d:E \rightarrow F$. The function $\displaystyle d$ induces a function 
$\displaystyle d^{*}: E^{*} \rightarrow F^{*}$ which transforms a word $\displaystyle \eta \in E^{*}$ into the word $\displaystyle d^{*}(\eta)$ which is obtained by replacing each letter $a$  of $\eta$ with the letter $d(a)$. 
The pair of functions  $\left[f,d\right]$ satisfies the property: for every $\displaystyle i \in \left\{1...n\right\}$ we have 
\begin{enumerate}
\item[(a)] $\displaystyle t{'}_{f(i)} = t_{i}$
\item[(b)] $\displaystyle \eta^{'}_{f(i)} = d^{*}\left(\eta_{i}\right)$
\item[(c)] $f$ is injective
\item[(d)] $d$ restricted to $Bound(P)$ takes  values in $Bound(Q)$ and is injective
\item[(e)] $d$ restricted to $Free(P)$ is at most 2-to-1.
\end{enumerate}
In this category of mol patterns, two of them are equivalent if they are isomorphic. 

Part of the condition (d) (that $d$ takes bound edge decorations to bound edge decorations) and the condition (e) are needed, even if they seem overly complex. In fact we shall identify molecules with decorated graphs as made of half-edges. The mentioned part of the condition (d) means just that the morphism preserves edges of these graphs. The condition (e) allows pairs of half-edges which are not part of edges in (the graph associated to) P to be transformed in pairs of half-edges which form an edge (in the graph associated to) Q. 

At the level of the library of programs, we have to be sure that our programs respect this equivalence relation. 

A mol pattern contains more information than a molecule, but this information is not geometrical and has to be destroyed. One of the means to keep only the geometrical information is to use random permutations (or morphisms) where needed in the algorithms. 

\subsection{Molecules as graphs}

A molecule can be turned into a graph. This means we need a conversion from a mol pattern to a graph such that two isomorphic mol patterns are converted into the same graph. In order to understand 
the conversion (done by functions in \href{https://github.com/mbuliga/quinegraphs/blob/master/js/ioprep.js}{ioprep.js}), we have to understand the relation between mol patterns and graphs. 

A graph is pair (G,E) where G is a finite nonempty set of nodes and E is a set of edges. Each edge is a set of two nodes (therefore the nodes have to be different). We say that an edge links the two nodes. There are no edges which connect a node with itself.

An oriented graph is a different mathematical object. It is a pair of functions (source, target) defined on a set E of edges, with values in a set G of nodes. An oriented graph admits edges with the source and target  being the same node. If an oriented graph does not have such edges, then it can be turned into a graph (as  defined previously) by associating to each edge e the set formed by the nodes source(e) and target(e). 

A graph can also be seen as a collection of half-edges, according to the following definition. A graph is a finite set of H of half-edges, with a set N of disjoint sets of half-edges, called nodes, and a set E of disjoint unordered pairs of half-edges, called edges. 

We shall use for our graphs the following modification of a graph as a collection of half-edges. Let C be a set of colors. A C-decorated graph is:
\begin{enumerate}
\item[-] a finite collection of half-edges H, 
\item[-] with a set N of nodes, where each node is a vector $\displaystyle [c,h_{1},...,h_{k}]$ where $k$ is the arity of the node and  $h_{1},...,h_{k}$ are half-edges, such that a half-edge $h \in H$ appears exactly once in one of the nodes,
\item[-] and with a set of unordered pairs of half-edges, called edges.
\end{enumerate}

We transform a C-decorated graph G into a unoriented usual graph G' with colored nodes. Denote by $m$ the maximal arity of nodes in G. Then we add to the set of colors the set $\displaystyle \left\{1,...,m\right\}$ (we suppose of course that $C$ is disjoint from this set) and we obtain a new set of colors $C'$. To each node  $\displaystyle n = [c,h_{1},...,h_{p}]$ in G, of arity p, we add p + 1 nodes  in G', the $k$-th node colored with $"k"$, for $k = 1,...,p$, and the last node colored with $"c"$. This last node is named a center node and the other nodes are named port nodes. We add edges in G' from the center node to each of its port nodes, we call these internal edges. Finally, for each pair of half-edges which form an edge in G, we add an external edge linking the corresponding node ports.

d3 graphs are oriented graphs, in the format used in the js library \href{https://d3js.org/}{d3.js}. See more about d3 graphs in \href{https://github.com/mbuliga/quinegraphs/blob/master/js/myD3Graph.js}{myD3Graph.js}. A d3 graph is given as a pair of vectors,  called nodes and links. 

The nodes vector has elements, each one (node) is an object
\begin{enumerate}
\item[]\begin{verbatim}
{"id": id, 
"type": type, 
x: x, y: y, 
vx:0, vy:0, 
links:[], 
"age":age}
\end{verbatim}
\end{enumerate}
where "id" is the identity (index of the node in the nodes vector), "type" is the graph node type, x, y, vx, vy  are the coordinates and velocities (used in the force graph simulation done by d3), links is a vector of links (i.e. edges) connected with the present node. "age" is the age of the node, used in some of the experiments. Possibly other fields may be added.

The links vector describes the edges of the oriented graph. Each links element is an object 
\begin{enumerate}
\item[]\begin{verbatim}
{"source": nsource, 
"target": ntarget, 
"value": value, 
"age":age}
\end{verbatim}
\end{enumerate}
where "source" is the node (index) of the source node, "target" is the node (index) of the target node and "value" is a number which is used for the width of the link (edge) as drawn by d3.js. Internal edges (those from the center node to its port nodes) and the external edges will have different width. The paramater "age" is the age of the link, used in some experiments. Possibly other fields may be added.

An oriented graph structure. like these d3 graphs, do contain non-geometrical information, like the one we mentioned previously concerning mol patterns. We have to be careful to not use this information, or to destroy it somehow, in the algorithms.

Here we use d3 graphs as usual graphs, in the sense that we don't treat edges as oriented. (At the level of the programs, for example starting with a node, we may search for the other nodes which are linked to this one, irrespective to the fact that they are sources or targets). This is done in the following way. 

\subsubsection{Graph nodes}

\paragraph{Graph nodes types.} We introduce a set GT of graph node types. Further we exploit the fact that here we use mol node types with valence at most 3 (but the extension to any valence should be obvious). 

If NT is the set of mol node types, then GT is obtained from NT by adding 3 new types: 
\begin{enumerate}
\item[-] "in", "middle", "out".
\end{enumerate}

We also have a predicate isCenter (defined in \href{https://github.com/mbuliga/quinegraphs/blob/master/js/myD3Graph.js}{myD3Graph.js}), which is true for any graph node type which is not "in", "middle" or "out". We  say that a graph node is a center if it has a graph node type for which isCenter is true. 

A node which is not a center is called a port node.

GT is kept in the vector graphNodes, in the script \href{https://github.com/mbuliga/quinegraphs/blob/master/js/nodes.js}{nodes.js}.

\begin{enumerate}
\item[]\begin{verbatim}
graphNodes = [
"in","out","middle",
"L","A","FI","D","FOE","FOX","FO",
"T","FRIN","FROUT","Arrow",
"GAMMA","DELTA"
];
\end{verbatim}
\end{enumerate}

Because we want to represent graphically the nodes, we associate colors to each graph node type. Remark that we don't use a bijection from the graph node types to the colors.

Each  graph node (which we are going to define from a mol pattern) is graphically represented by a colored circle. The radius of the circle is a function of the predicate isCenter, that is center nodes have a radius and the port nodes have a different radius.

\paragraph{From mol nodes to graph nodes.} To each mol node we associate a GT-decorated graph. The association is the following. 

There is a center node according to the mol node type. There are port nodes, one for each letter of the mol node word. Here we exploit the fact that we use mol nodes of valence at most 3. 

The function which associates a vector of graph node types (among "in", "out" or "middle") to the valence of a mol node is  nodePortTypes, in the script \href{https://github.com/mbuliga/quinegraphs/blob/master/js/nodes.js}{nodes.js}.
\begin{enumerate}
\item[]\begin{verbatim}
nodePortTypes = [
["in"],
["in","out"],
["in","middle","out"]
];
\end{verbatim}
\end{enumerate}

So the first port node is always "in", the last one (for valence at least 2) is always "out", the remaining port is "middle".

The graph associated to a mol node is formed by the center node and the port nodes, with edges from the center nodes to the port nodes. 
Specifically, we draw a center graph node as an "o" and a port node as a dash "|" with a number (starting from 1). 

\begin{figure}[h!]
\centering
  \includegraphics[width=0.25\linewidth]{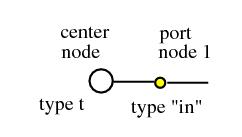}
  \caption{The graph associated to a 1-valent mol node}
  \label{fig:valence-1}
\end{figure}

Then a mol node of valence 1, of mol node type "t" and edge "a" 
\begin{enumerate}
\item[]\begin{verbatim}
t a
\end{verbatim}
\end{enumerate}
becomes a graph as in figure \ref{fig:valence-1}.

\begin{figure}[h!]
\centering
  \includegraphics[width=0.35\linewidth]{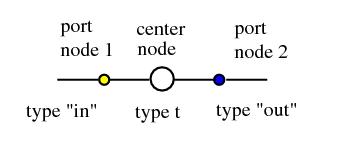}
  \caption{The graph associated to a 2-valent mol node}
  \label{fig:valence-2}
\end{figure}

\begin{figure}[h!]
\centering
  \includegraphics[width=0.4\linewidth]{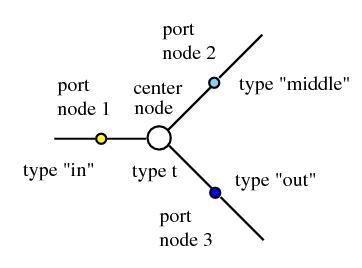}
  \caption{The graph associated to a 3-valent mol node}
  \label{fig:valence-3}
\end{figure}

A mol node of valence 2, of mol node type "t" and edges "ab"
\begin{enumerate}
\item[]\begin{verbatim}
t a b
\end{verbatim}
\end{enumerate}
becomes a graph as in figure \ref{fig:valence-2}.
A mol node of valence 3, of mol type "t" and edges "abc"
\begin{enumerate}
\item[]\begin{verbatim}
t a b c
\end{verbatim}
\end{enumerate}
becomes a graph as in figure \ref{fig:valence-3}.

\subsubsection{Mol patterns as graphs}
To a mol pattern we associate the graph formed by (the graphs of) all mol nodes, with edges connecting port nodes which have the same edge tag. The definition of a mol pattern assures us that this is possible, because each edge tag cannot occur more than twice. 

Let us come back to the d3 graphs, which are oriented. The orientation (i.e. which node is the source and which one is the target) will be neglected. This is possible because the graphs which we obtain do not have edges which connect a node with itself. Indeed, even if an edge tag appears twice in the same mol node, the corresponding edge connects different node ports. 

\section{Rewrites}
\label{rewrites-prog}

A rewrite is an object, like 
\begin{enumerate}
\item[]\begin{verbatim}
{
left:"A",right:"FO",             \\ the nodes types of the LHS pattern
action:"DIST1",                  \\ the action name
named:"A-FO",                    \\ the name of the rewrite
t1:"FOE",t2:"FOE",t3:"A",t4:"A", \\ the node types of the RHS pattern
blocks:["FOE-A"],                \\ patterns which have LHS in this RHS
kind:"DIST"                      \\ the general type of rewrite
}
\end{verbatim}
\end{enumerate}

\subsection{Rewrites of mol patterns}

A mol pattern rewrite is a \href{http://eprints.adm.unipi.it/1950/1/TR-96-17.ps.gz}{double push-out} \cite{dpo} rewrite. In this case of mol patterns, a general rewrite is defined by a pair of mol 
patterns $[LHS,RHS]$, with the property that 
$$Free(LHS) = Free(RHS)$$
$LHS$ is called the left hand side pattern and $RHS$ is called the right hand side pattern. $\displaystyle Free(LHS) = Free(RHS)$ is the interface.

Given a mol pattern $M$ and a rewrite $[LHS,RHS]$, an application of the rewrite consists of: 
\begin{enumerate}
\item[-] a morphism from $LHS$ to $M$, with image $\displaystyle match^{L}(LHS)$
\item[-] a morphism from $RHS$ to $M$, with image $\displaystyle match^{R}(RHS)$, such that 
\begin{enumerate}
\item[(var1)] $\displaystyle match^{R}(Free(RHS)) = match^{L}(Free(LHS))$
\item[(var2)] $\displaystyle match^{R}(Bound(RHS)) \cap Bound(M) = \emptyset$
\end{enumerate}  
\item[-] from the mol pattern M are eliminated (spliced) the mol nodes of $\displaystyle match^{L}(LHS)$
\item[-] the mol modes of $\displaystyle match^{R}(RHS)$ are added to the result.
\end{enumerate}

This definition of a rewrite application has to be backed by algorithms which can be implemented by local machines (section \ref{localmachines}):
\begin{enumerate}
\item[(a)] from $M$ and $LHS$ output  $\displaystyle match^{L}(LHS)$
\item[(b)] from $M$, $LHS$, $RHS$ and $\displaystyle match^{L}(LHS)$ output $\displaystyle match^{R}(RHS)$
\item[(c)] from $M$, $\displaystyle match^{L}(LHS)$ and $\displaystyle match^{R}(RHS)$ output $M'$, the mol pattern after the rewrite.
\end{enumerate}
Among these algorithm, (c) is clear, (a) is a matching algorithm  and (b) is more problematic because of the condition (var2), which 
asks for a way to produce new bound edges, i.e. new names which are not in $Bound(M)$. As (c) is trivial, we may think about it as the final part 
of the algorithm (b), thus we have: 
\begin{enumerate}
\item[(a)] match algorithm: from $M$ and $LHS$ output  $\displaystyle match^{L}(LHS)$
\item[(b)] rewrite algorithm: from $M$, $LHS$, $RHS$ and $\displaystyle match^{L}(LHS)$ output $\displaystyle match^{R}(RHS)$,then replace $\displaystyle match^{L}(LHS)$ by $\displaystyle match^{R}(RHS)$ in $M$. Output $M'$, the mol pattern after the rewrite.
\end{enumerate}

It is also interesting to consider a finite collection of general rewrites $$\displaystyle [LHS_{1},RHS_{1}], ...,[LHS_{N},RHS_{N}]$$ 
Algorithms $\displaystyle (a)_{k}$ and $\displaystyle (b)_{k}$ for all $k = 1...N$ can be merged into one (a) algorithm and (b) algorithm. 
This can be done by abstracting over the node types. Looking back at the definition of molecules as equivalence classes of mol patterns, 
we could consider a larger equivalence relation, by admitting mol pattern morphisms which do not preserve mol node types. Instead we might 
treat mol node types as we treated edges tags, perhaps retaining only nodeValence of the node type (or only the nodes arities). 

The most general formalism may turn out to be much more verbose than what we need. That is why let's make only the following definition: a connection morphism between two mol patterns
\begin{enumerate}
\item[] $\displaystyle P = \left[(t_{i}, \eta_{i})\right]_{i = 1...n}$, which is E-decorated, with mol node types in NT,
\item[] $\displaystyle Q = \left[(t{'}_{j}, \eta^{'}_{j})\right]_{j = 1...m}$, which is F-decorated, with mol node types in NT',
\end{enumerate}
 is a triple of functions $\left[f,d, g\right]$, where $\displaystyle f:\left\{1...n\right\} \rightarrow \left\{1...m\right\}$, $\displaystyle d:E \rightarrow F$ and $\displaystyle g:NT \rightarrow NT'$ with  the properties: for every $\displaystyle i \in \left\{1...n\right\}$ we have 
\begin{enumerate}
\item[(a)] $\displaystyle t{'}_{f(i)} = g(t_{i})$
\item[(b)] $\displaystyle \eta^{'}_{f(i)} = d^{*}\left(\eta_{i}\right)$
\item[(c)] $f$ and $g$ are injective
\item[(d)] $d$ restricted to $Bound(P)$ takes  values in $Bound(Q)$ and is injective
\item[(e)] $d$ restricted to $Free(P)$ is at most 2-to-1.
\end{enumerate}
A connection pattern is then an equivalence class of mol patterns up to connection isomorphisms. 

Connection patterns rewrites are defined like mol patterns rewrites. The advantage is that we may compress a finite collection of 
general mol pattern rewrites into a much smaller collection of connection patterns rewrites. For a general connection pattern rewrite we shall use the name "action" (which explains the field "action" in the description of a rewrite).

At the action level, all LHS patterns we shall consider are of the form.
\begin{enumerate}
\item[]\begin{verbatim}
n1type e ...
n2type ...e...
\end{verbatim}
\end{enumerate}
where  n1type and n2type are the types of the nodes. To identify such a LHS pattern we need to specify 
\begin{enumerate}
\item[]\begin{verbatim}
left:n2type, right:n1type,    \\ the two nodes types of the LHS pattern
named:n2type + "-" + n1type,  \\ the name of the rewrite
\end{verbatim}
\end{enumerate}
Here the "left" and "right" node type come from the image that the edge "e" is oriented from left to right, with the mol node of type n1type at right. 

This notation for a LHS pattern supposes that there is only one connection pattern for a pair of node types n1type, n2type. The connection pattern will appear as a predicate in the algorithm (a). 

Each action (i.e. connection pattern rewrite) is identified by a name, for our example: 
\begin{enumerate}
\item[]\begin{verbatim}
action:"DIST1",                  \\ the action name
\end{verbatim}
\end{enumerate}

Finally, the RHS patterns are also given via their connection patterns, for our example: 
\begin{enumerate}
\item[]\begin{verbatim}
t1:"FOE",t2:"FOE",t3:"A",t4:"A", \\ the node types of the RHS pattern
\end{verbatim}
\end{enumerate}

The algorithm (b) takes as input:
\begin{enumerate}
\item[-] the $LHS$ mol pattern $\displaystyle match^{L}(LHS)$ , given by the algorithm (a)
\item[-] the action name (which gives the connection pattern rewrite) and the node types needed to build the $\displaystyle match^{R}(LHS)$ pattern. 
\end{enumerate}
The problem of how to generate new names which are not in $Bound(M)$ remains. 

\subsection{Graph rewrites}

At the graph level, the connection pattern
\begin{enumerate}
\item[]\begin{verbatim}
n1type e ...
n2type ...e...
\end{verbatim}
\end{enumerate}
translates into a center node of type $n1type$, whose first node port is connected to a node port of a center node of type $n2type$. The match algorithm (a) is the function findTransform(n1) from \href{https://github.com/mbuliga/quinegraphs/blob/master/js/chemistry.js}{chemistry.js}, where n1 is the center node of type  $n1type$. The function findAllTransforms() uses the previous function for all center nodes to make a vector of all possible graph rewrites.

The algorithm (b) is embodied by the function doTransform(n1, trans)from \href{https://github.com/mbuliga/quinegraphs/blob/master/js/chemistry.js}{chemistry.js}, which takes as input the center node n1 and a matching graph rewrite trans. Instead of generating new names (of nodes or edges), the algorithm delegates this to d3.js part, by using functions like addNodeAndEdges, or removeNodeAndEdges, which are defined in \href{https://github.com/mbuliga/quinegraphs/blob/master/js/myD3Graph.js}{myD3Graph.js}. 

\subsection{A solution of the problem of new names}
\label{hapax}

\href{https://github.com/mbuliga/hapax}{Project Hapax} gives such a solution, described \href{https://mbuliga.github.io/hapax/chemlambda-and-hapax.html}{here}. 

The idea is simple: we can exclude the need of new names if we reformulate all rewrites as permutations of edges. As an example, let's consider 
the rewrite "L-A", as it appears in chemlambda (and which is a well known graphical version of the $\beta$ rewrite). In terms of $LHS, RHS$ mol patterns, this rewrite transforms the $LHS$ 
\begin{enumerate}
\item[]\begin{verbatim}
L c b a
A a d e
\end{verbatim}
\end{enumerate}
into the $RHS$
\begin{enumerate}
\item[]\begin{verbatim}
Arrow c e
Arrow d b
\end{verbatim}
\end{enumerate}
(Here we don't need new edge names, but the example is easy to follow.) 

This rewrite can be reformulated as: transform the new $LHS'$ pattern
\begin{enumerate}
\item[]\begin{verbatim}
L c b a
A a d e
Arrow b' a'
Arrow a' b'
\end{verbatim}
\end{enumerate}
into the new $RHS'$ pattern
\begin{enumerate}
\item[]\begin{verbatim}
L b' a b'
A a' a a'
Arrow c e
Arrow d b
\end{verbatim}
\end{enumerate}
This can be further restated as a chemical reaction: 
\begin{enumerate}
\item[]\begin{verbatim}
L c b a       Arrow b' a'        Arrow c e       L b' a b'
A a d e   +   Arrow a' b'   -->  Arrow d b   +   A a' a a'
\end{verbatim}
\end{enumerate}
which has the form 
\begin{enumerate}
\item[]\begin{verbatim}
LHS   +   Token1   -->  RHS   +   Token2
\end{verbatim}
\end{enumerate}
Here $Token1$ and $Token2$ are 2-mol nodes patterns which have to be added to the rewrite in order to make it conservative. 

The solution thus transfers the problem of generating new names to the problem of generating new tokens. Or, we know how to reliably generate new tokens. We may imagine that $Token1$, $Token2$ are money tokens and that each rewrite involves a cost (but mind that all it really matters is that there exist well known ways to generate new tokens in a decentralized way).

\end{document}